\documentclass[journal]{IEEEtran}

\usepackage{CJKutf8}
\usepackage{cite}
\usepackage{color,xcolor}
\usepackage{amsmath,amssymb,amsfonts}
\usepackage{algorithm}
\usepackage{algpseudocode}
\usepackage{graphicx}
\usepackage{textcomp}
\usepackage{wrapfig}
\usepackage{epstopdf}
\usepackage{textcomp}
\usepackage{array}
\usepackage{stfloats}
\usepackage{url}
\usepackage{booktabs}
\usepackage{threeparttable}
\usepackage{multirow}
\usepackage{bm}
\usepackage{gensymb}
\usepackage{graphics} 
\usepackage{epsfig} 
\usepackage{times} 
\usepackage{balance}
\usepackage{stfloats}
\usepackage{setspace}

\title{\LARGE \bf Split Covariance Intersection Filter Based Visual Localization With Accurate AprilTag Map For Warehouse Robot Navigation}
\author{Susu~Fang$^{1}$, Yanhao~Li$^{2}$, Hao~Li$^{1}$ 
	\thanks{$^{1}$Department of Automation, Shanghai Jiao Tong University, Shanghai, 200240, China.}
	\thanks{$^{2}$Université Paris-Saclay, ENS Paris-Saclay, CNRS, Centre Borelli, 91190, Gif-sur-Yvette, France.}
	\thanks{Corresponding author: Hao Li (email: haoli@sjtu.edu.cn).}
}

\begin{document}
\maketitle
\thispagestyle{empty}
\pagestyle{empty}

\begin{abstract}
Accurate and efficient localization with conveniently-established map is the fundamental requirement for mobile robot operation in warehouse environments. An accurate AprilTag map can be conveniently established with the help of LiDAR-based SLAM. It is true that a LiDAR-based system is usually not commercially competitive in contrast with a vision-based system, yet fortunately for warehouse applications, only a single LiDAR-based SLAM system is needed to establish an accurate AprilTag map, whereas a large amount of visual localization systems can share this established AprilTag map for their own operations. Therefore, the cost of a LiDAR-based SLAM system is actually shared by the large amount of visual localization systems, and turns to be acceptable and even negligible for practical warehouse applications.
Once an accurate AprilTag map is available, visual localization is realized as recursive estimation that fuses AprilTag measurements (i.e. AprilTag detection results) and robot motion data. AprilTag measurements may be nonlinear partial measurements; this can be handled by the well-known extended Kalman filter (EKF) in the spirit of local linearization. AprilTag measurements tend to have temporal correlation as well; however, this cannot be reasonably handled by the EKF. The split covariance intersection filter (Split CIF) is adopted to handle temporal correlation among AprilTag measurements. The Split CIF (in the spirit of local linearization) can also handle AprilTag nonlinear partial measurements. The Split CIF based visual localization system incorporates a measurement adaptive mechanism to handle outliers in AprilTag measurements and adopts a dynamic initialization mechanism to address the kidnapping problem. A comparative study in real warehouse environments demonstrates the potential and advantage of the Split CIF based visual localization solution.
\end{abstract}

\section{INTRODUCTION}
Robot localization is fundamental for mobile robotics and is involved in a large variety of practical applications \cite{durrant2006simultaneous, bailey2006simultaneous, vasiljevic2016high, bresson2017simultaneous, yu2013visual, li2019deep, xu2021indoor, Li2024IV}. Visual localization plays an important role in mobile robotics, thanks to its commercial competitiveness. It is true that a visual localization system may be susceptible to light conditions and has comparatively short perception range in contrast with LiDAR-based localization which also plays an important role in mobile robotics. However, these limitations of visual localization are naturally overcome in the context of indoor mobile robotics such as warehouse applications, because in indoor environments artificial light conditions are usually stable and the need for long-range perception (such as that in outdoor applications) is rare.

Visual simultaneous localization and mapping (SLAM) \cite{mur2015orb, sumikura2019openvslam, campos2021orb, bloesch2015robust, qin2018vins, Li2024arXivVisualSLAMMOT} is a special form of visual localization, usually without \textit{a priori} map. Despite its popularity and its merits for flexible exploration in an unknown environment, visual SLAM without \textit{a priori} map is unlikely to be a proper solution for a known environment of which an accurate map can be established. After all, an accurate map can largely facilitate visual localization.

How to conveniently establish an accurate map is a concern for many researchers. Some researchers rely on artificial markers, such as binary BCH code\cite{zheng2008mr}, fiducial markers\cite{fourmy2019absolute}, and AprilTag \cite{olson2011apriltag, wang2016apriltag}. The AprilTag, which is illustrated in blue comment boxes of Fig. \ref{Fig.1}(a) and (b), is a kind of popular and commonly-used artificial marker, thanks to the ease of its deployment and to the richness of information that it can convey. We also adopt the AprilTag and base the intended visual localization on an accurate AprilTag map established \textit{a priori}.

An AprilTag map may be established in the visual SLAM way \cite{pfrommer2019tagslam}. However, considering the natural advantage of LiDAR-based SLAM systems over visual SLAM systems in terms of accuracy and robustness, we choose to establish an intended AprilTag map of the warehouse environment with the help of LiDAR-based SLAM. It is true that a LiDAR-based system is usually not commercially competitive in contrast with a vision-based system, yet fortunately for warehouse applications, only a single LiDAR-based SLAM system is needed to establish an accurate AprilTag map, whereas a large amount of visual localization systems can share this established AprilTag map for their own operations. Therefore, the cost of a LiDAR-based SLAM system is actually shared by the large amount of visual localization systems, and turns to be acceptable and even negligible for practical warehouse applications.

Once an accurate AprilTag map is available, visual localization is realized as recursive estimation that fuses AprilTag measurements (i.e. AprilTag detection results) and robot motion data \cite{kayhani2019improved, yu2021indoor, hanley2021impact}. 

AprilTag measurements may be nonlinear partial measurements. For example, the performance of the AprilTag detection module deteriorates as the view distance and the view angle increase \cite{li2014mobile, ullah2020simultaneous}. Sometimes the AprilTag detection module may output distance measurements only instead of complete pose measurements \cite{pfrommer2019tagslam}. This is likely to happen when its sub-step of homography computation encounters the view singularity problem due to visual data errors and noises. Nonlinear partial measurements can be handled by the well-known extended Kalman filter (EKF) in the spirit of local linearization \cite{kayhani2019improved} \cite{Li2022FARET_en} \cite{li2022FARET}. 

AprilTag measurements tend to have (unknown) temporal correlation; however, this cannot be reasonably handled by the EKF. The split covariance intersection filter (Split CIF) \cite{Julier2001} \cite{li2013split} \cite{Cros2025} is adopted to handle temporal correlation among AprilTag measurements --- The Split CIF, which may be regarded as a generalization of both the (extended) Kalman filter and the covariance intersection filter, can reasonably handle both known independent information and unknown correlated information in source data. It has been applied in a number of intelligent vehicle applications \cite{li2013cooperative, Li2013d, Wanasinghe2014, Pierre2018, Chen2020, fang2022inertial, Allig2022, Li2022TITS}.

AprilTag measurements may have outliers as well. Discarding abnormal data whenever found is an easy choice, yet this may result in loss of some ``not so bad'' and exploitable information \cite{zair2016outlier, bai2018robust}. Inspired by the Gauss-Newton iterative methods \cite{shojaei2011experimental, zhao2016robust} and the innovation-based adaptive estimation methods \cite{huang2017new, wang2019adaptive}, a measurement adaptive mechanism is incorporated to handle outliers in AprilTag measurements.
Besides, the robot may accidentally encounter the kidnapping problem and this is handled by a dynamic initialization mechanism.

The solution of split covariance intersection filter based visual localization with accurate AprilTag map is proposed for warehouse robot navigation. It is realized as recursive estimation that fuses AprilTag measurements and robot motion data. It can naturally handle temporal correlation among AprilTag measurements and can handle AprilTag nonlinear partial measurements as well. It incorporates a measurement adaptive mechanism to handle outliers in AprilTag measurements and adopts a dynamic initialization mechanism to address the kidnapping problem. A comparative study in real warehouse environments is presented to demonstrate the potential and advantage of the proposed solution of the Split CIF based visual localization with accurate AprilTag map.

\section{AprilTag Map Establishment}

Visual simultaneous localization and mapping (SLAM) \cite{mur2015orb, sumikura2019openvslam, campos2021orb, bloesch2015robust, qin2018vins} is nowadays popular and has its merits for flexible exploration in an unknown environment. However, visual SLAM without \textit{a priori} map is unlikely to be a proper solution for a known environment of which an accurate map can be established. After all, an accurate map can largely facilitate visual localization.

LiDAR-based SLAM systems \cite{shan2018lego} possess natural advantage over visual SLAM systems \cite{pfrommer2019tagslam} in terms of accuracy and robustness, especially for outdoor application environments such as traffic environments and large indoor application environments such as warehouse environments. So we choose to establish an intended AprilTag map of the warehouse environment with the help of LiDAR-based SLAM. More specifically, we rely on LiDAR-based SLAM to obtain accurate estimation of robot ego-poses during AprilTag map establishment, such that \textit{a priori} registration of AprilTags have ``good anchors'' for visual mapping. 

\begin{figure}[htbp]
	\centering
	\includegraphics[scale=0.50]{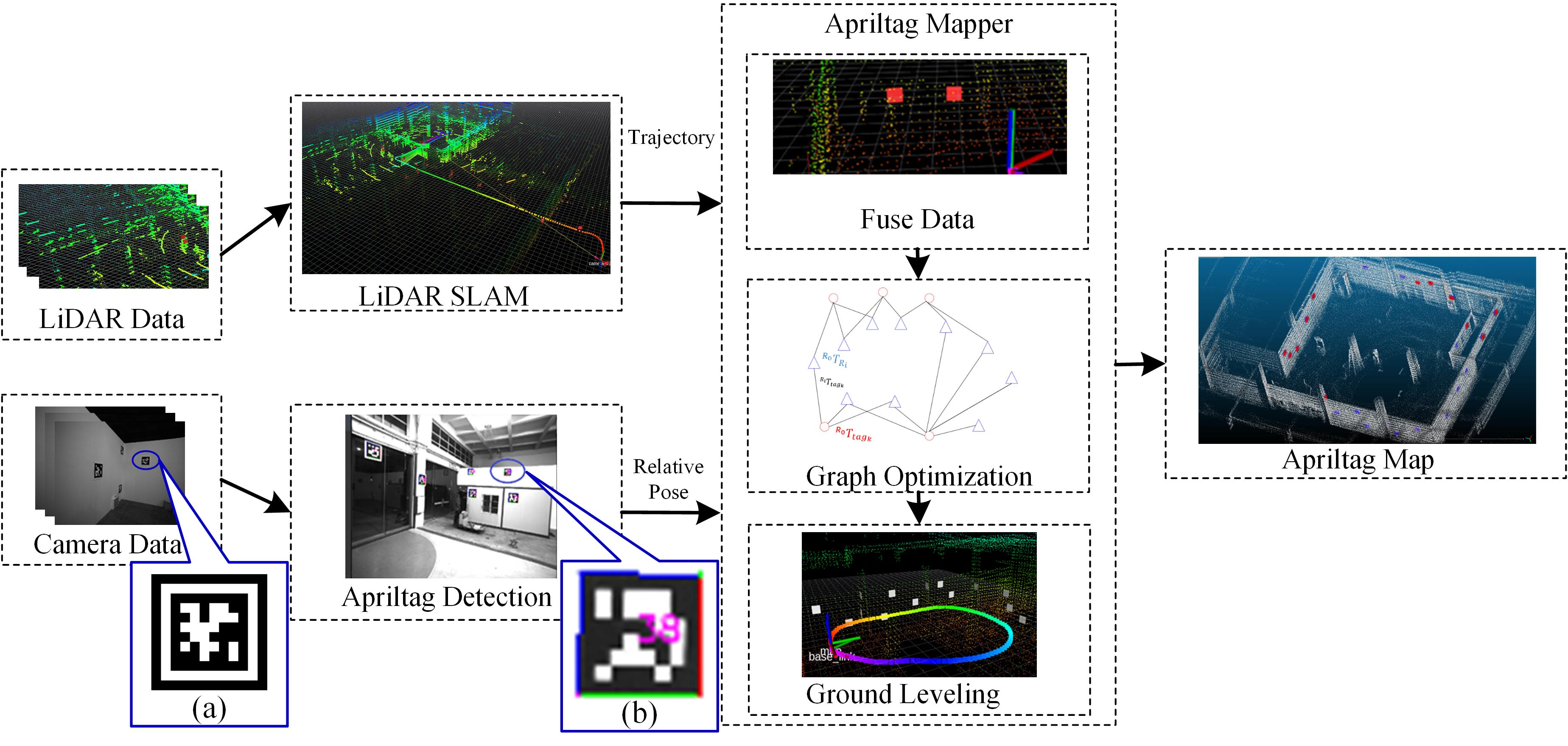}    		
	\caption{AprilTag map establishment (The AprilTag is shown in (a) and the AprilTag after detection and recognition is shown in (b).)}\label{Fig.1}
\end{figure}

The process of AprilTag map establishment using LiDAR-based SLAM is illustrated in Fig. \ref{Fig.1}. For the LiDAR-equipped robot deployed for sake of AprilTag mapping (we call it the mapping robot), camera-LiDAR co-calibration can be done \textit{a priori} \cite{Li2013calib}. 

Equipped with the LiDAR and camera, the mapping robot can drive within the AprilTag's visible area and record LiDAR point cloud and camera image data while driving. We adapt the LiDAR SLAM method to acquire a trajectory of the robot, which can be used as the ground truth. At the same time, we know the calibration parameters of the camera and robot, as well as the AprilTag image data at every moment by AprilTag detection. Then, we can solve the relative pose of each AprilTag relative to the robot at each moment. Finally, the graph optimization module to solve the relative pose among each AprilTag, and transform the AprilTag pose to the ground plane coordinate system to generate the map.

Specifically, the LiDAR SLAM module can be any representative LiDAR SLAM framework, and here we use the lightweight and ground-optimized LOAM (LeGO-LOAM) \cite{shan2018lego} for its superior performance with limited resources. The Apriltag3 \footnote{https://github.com/AprilRobotics/apriltag}\footnote{http://wiki.ros.org/apriltag\_ros}algorithm is adapted for the AprilTag detection module due to the improved performance and detection efficiency compared with the previous version. It can also solve the relative pose of the AprilTag in the camera coordinate system. For the graph optimization module, the pose of the robot at each moment ${^{R_0}{\bf T}} _{R_k}$ is known and fixed. ${^{R_0}{\bf T}} _{tag_m}$ is the item to be optimized, which denotes the pose of observed AprilTag  \{${tag_m}, m \in M_k$\}, where $M_k$ is the ID set of the Apriltags observed at $k$ relative to the robot coordinates system $R$. We use the GTSAM \cite{dellaert2012factor} optimizer to solve such pose graph optimization problem that find the optimal pose of all Apriltags relative to the robot coordinate system at the starting moment $\{{^{R_0}{\bf T}} _{tag_m}, m \in M\}$. The nodes, factors and constraints of the graph optimization process are the ${^{R_0}{\bf T}} _{tag_m}$, the relative pose of the robot coordinate system at each time $R_k$ relative to that at the starting moment ${^{R_0}{\bf T}} _{R_k}$ and the pose of Apriltags relative to the robot coordinate system at each moment ${^{R_k}{\bf T}} _{tag_m}$ respectively, which also can be seen in the graph optimization module in Fig. \ref{Fig.1}.  

It is true that a LiDAR-based system is usually not commercially competitive in contrast with a vision-based system, yet fortunately for warehouse applications, only a single LiDAR-based SLAM system (namely the unique mapping robot) is needed to establish an accurate AprilTag map, whereas a large amount of visual localization systems (namely dozens and even hundreds of warehouse robots that actually operate) can share this established AprilTag map for their own operations. Therefore, the cost of a LiDAR-based SLAM system is actually shared by the large amount of visual localization systems, and turns to be acceptable and even negligible for practical warehouse applications.

\section{Split Covariance Intersection Filter Based Visual Localization}

\subsection{Split Covariance Intersection Filter}
The detailed derivation of split covariance intersection filter (Split CIF) theory has been shown in \cite{li2013split}. In real implementations, for two estimations to-be fused: \{${{\bf{X}}_1},{{{\bf{P}}_{1}}}$\} and \{${{\bf{X}}_2},{{{\bf{P}}_{2}}}$\}, where the $\bf X$ and $\bf P$ are the estimate state and covariance, respectively. ${{\bf{X}}_1}$ is supposed to complete observation of the true state ${{\bf{X}}_{true}}$ in general i.e., ${{\bf{X}}_1} = {{\bf{X}}_{true}}$, whereas the ${{\bf{X}}_2}$ is the complete or partial observation, which is generally denoted as ${{\bf{X}}_2} = {{\bf{H}}}{{\bf{X}}_{true}}$,  where ${\bf{H}}$ denotes the measurement matrix. The Split CIF decomposes the covariance into the correlated component (subscripts $d$) and the independent component (subscripts $i$). Therefore,  the two data sources \{${{\bf{X}}_1},{{{\bf{P}}_{1i}}+{{\bf{P}}_{1d}}}$\} and \{${{\bf{X}}_2},{{{\bf{P}}_{2i}}+{{\bf{P}}_{2d}}}$\} can be fused by the Split CIF formula as follows:
\begin{equation}
\begin{small}
\begin{aligned}
{{\bf{P}}_1} &= {{\bf{P}}_{1d}}/{\omega _{opt}} + {{\bf{P}}_{1i}}\\
{{\bf{P}}_2} &= {{\bf{P}}_{2d}}/(1 - {\omega _{opt}}) + {{\bf{P}}_{2i}}\\
{\bf{K}} &= {{\bf{P}}_1}{{\bf{H}}^T}{({\bf{H}}{{\bf{P}}_1}{{\bf{H}}^T} + {{\bf{P}}_2})^{ - 1}}\\
{\bf{X}} &= {{\bf{X}}_1} + {\bf{K}}({{\bf{X}}_2} - {\bf{H}}{{\bf{X}}_1})\\
{\bf{P}} &= ({\bf{I}} - {\bf{KH}}){{\bf{P}}_1}\\
{{\bf{P}}_i} &= ({\bf{I}} - {\bf{KH}}){{\bf{P}}_{1i}}{({\bf{I}} - {\bf{KH}})}^T + {\bf{K}}{{\bf{P}}_{2i}}{{\bf{K}}^T}\\
{{\bf{P}}_d} &= {\bf{P}} - {{\bf{P}}_i}
\end{aligned}
\end{small}
\end{equation}
Where ${\omega _{opt}} \in [0,1]$, and ${\omega _{opt}}$ is determined by solving a convex optimization problem (see \cite{Li2022FARET_en} \cite{li2022FARET} for details). In addition, about the Split CIF, it can be regarded as a generalization of the Kalman filter, as shown in above equation. Specially, let ${{\bf{P}}_{1d}}$ and ${{\bf{P}}_{2d}}$ be zero, and above equation will become similar to the Kalman filter. 

Concerning the instability of the AprilTag detection module, the quality of measurements will be different and there exists many low accurate observations that can be regarded as outliers. We incorporate a measurement adaptive mechanism to handle outliers in AprilTag measurements. For these different quality observations, we process them with different way, which also can be called soft abandon mechanism. It is assumed that the predicted value is ${{\widetilde{\bf{X}}}_{k+1/k}}$, and then the current measurements is ${\bf{Z}}_{{k + 1}}$. When $||{{\bf{Z}}_{k + 1}}-{{\widetilde{\bf{X}}}_{k+1/k}}||$ is greater than a screening threshold, ${\bf{Z}}_{{k + 1}}$ is discarded. When observations are not discarded but are of lower quality, we drive an adaptive and accurate observation noise model which fits the relationship of observation error and view distance (denote as $L$) and view angle (denote as $\alpha$) according to experimental error statistics to dynamically evaluate the observation uncertainty by its noise covariance as follows: 
\begin{equation}
{\bf{R}}_{k+1}=0.25(L/(\alpha^2))||{{\widetilde{\bf{X}}}_{k+1/k}}-{{\bf{Z}}_{k + 1}}||
\end{equation}

Herein, the greater of the deviation between observation and the prediction state, the smaller the weight in the fusion, and its final impact on the fusion result is always limited. Compare with some Gauss-Newton iterative method or innovation-based method, such adaptive noise model is more flexibility and can fundamentally suppress the influence of outliers to a greater extent. Also, even though the measurements exceed the screening threshold, they are not discarded directly, and may still function properly without causing a complete waste of information. Of course, if the observation is very unreasonable, in order to avoid unreasonable errors, we will directly discard it. 

Implementation of the Split CIF that incorporates the measurement adaptive mechanism in the context of AprilTag-based visual localization is given as pseudo code in Alg. \ref{Alg:ImplSplitCIF}. In this algorithm, ${\bf{X}}, {{\bf{P}}_{i}}, {{\bf{P}}_{d}}$ denote the state vector and its independent and dependent covariance matrix in different estimation processes about time $k$. The Split CIF implementation uses the state evolution model i.e. the system model to obtain the prediction state with its split covariance \{${{\widetilde{\bf{X}}}_{{k+1/k}}}, {{\bf P}_{{i,k+1/k}}}+{{\bf P}_{{d,k+1/k}}}$\}, as shown in lines 1-3 in the prediction part of the following algorithm. ${{\bf{u}}_k}$ is the input control vector. ${{\bf{G}}_{x_{k}}}$ and ${{\bf{G}}_{u_{k}}}$ are the Jacobian matrices of the state evolution model $g(.)$ with respect to the state vector ${{\bf{X}}_{k}}$ and the control variable ${{\bf{u}}_k}$, respectively. ${{\bf{Q}}_{k}}$ is the known covariance of the process noise ${\bf{w}}_k$. ${{\bf{P}}_{pre,i,k}}$ denotes the prediction model error because this predictive model is a simplified vehicle model. For the pose of the AprilTag with identify information (ID)) relative to the camera obtained from the AprilTag detection module, we use the ID to perform map matching to obtain the global pose of the AprilTag, then use the calibration information of camera and robot to obtain the 6 DoFs global pose of the robot at each moment. For this 6 DoFs pose provided by AprilTag detection module, we use the ${{x_{msr}},{y_{msr}},{\theta_{msr}}}$ as new measurements in ${\bf{Z}}_{{k + 1}}$ with its covariance in split form \{${\bf{R}}_{{i,k+1}}+{\bf{R}}_{{d,k+1}}$\} of the observation noise  ${{\bf{v}}_{k+1}}$ to update the current obtained from prediction process. ${{\bf{H}}_{k+1}}$ is the Jacobian matrix of the observation model $h(.)$ in \{${{{\bf{Z}}_{k + 1}} = h({{\bf{X}}_{k+1/k}})+{{\bf{v}}_{k+1}}}$\} with respect to the state vector ${{\bf{X}}_{k+1/k}}$.

\begin{algorithm}
	\begin{spacing}{0.99}
		\caption{Implementation of the Split CIF for AprilTag-based visual localization}
		\label{Alg:ImplSplitCIF}
		$\ast$ \textbf{Prediction:}
		\begin{algorithmic}[1]
			\State\ ${{\widetilde{\bf{X}}}_{k + 1/k}} = g({{\widetilde{\bf{X}}}_{k}},{{\bf{u}}_k})$\\
			\ ${{\bf{P}}_{i,k + 1/k}} = {{\bf{G}}_{x_{k}}}{{\bf{P}}_{i,k}}{\bf{G}}_{x_{k}}^T + {{\bf{G}}_{u_{k}}}{{\bf{Q}}_{k}}{\bf{G}}_{u_{k}}^T+{{\bf{P}}_{pre,i,k}}$\\
			\ ${{\bf{P}}_{d,k + 1/k}} = {{\bf{G}}_{x_{k}}}{{\bf{P}}_{d,k}}{\bf{G}}_{x_{k}}^T$
		\end{algorithmic}
		$\ast$ \textbf{Update incorporating the measurement adaptive mechanism:}
		\begin{algorithmic}[1]
			\State	\ ${\bf{P}}_{1,k + 1/k} = {{\bf{P}}_{d,k + 1/k}}/{\omega _{opt}} + {{\bf{P}}_{i,k + 1/k}}$\\
			\ ${\bf{P}}_{2,k + 1} = {{\bf{R}}_{d,k + 1}}/({1-\omega _{opt}}) + {{\bf{R}}_{i,k + 1}}$\\		
			\ \begin{small}${\bf{K}}_{{k + 1}}\!\! = {\bf{P}}_{{1,k + 1/k}}{{\bf{H}}_{{k + 1}}}^T{({\bf{H}}_{{k + 1}}{\bf{P}}_{{1,k + 1/k}}{{\bf{H}}_{{k + 1}}}^T + {\bf{P}}_{2,k + 1})^{ - 1}}$\end{small}\\
			\ ${\widetilde{\bf{X}}}_{{k + 1}} = {\widetilde{\bf{X}}}_{{k + 1/k}} + {\bf{K}}_{{k + 1}}({{\bf{Z}}_{k + 1}} - {{\bf{H}}_{k+1}}{\widetilde{\bf{X}}}_{{k + 1/k}})$\\
			\ ${\bf{P}}_{{k + 1}} = ({\bf{I}} - {\bf{K}}_{{k + 1}}{\bf{H}}_{{k + 1}}){\bf{P}}_{{1,k + 1/k}}$\\
			\ $\begin{aligned}
			{\bf{P}}_{{i,k + 1}} &= ({\bf{I}} - {\bf{K}}_{{k + 1}}{\bf{H}}_{{k + 1}}){\bf{P}}_{{i,k + 1/k}}{({\bf{I}} - {\bf{K}}_{{k + 1}}{\bf{H}}_{{k + 1}})}^T\\&+{\bf{K}}_{{k + 1}}{\bf{R}}_{{i,k + 1}}{{{\bf{K}}_{{k + 1}}}^T}
			\end{aligned}$\\
			\ ${\bf{P}}_{d,k + 1}={\bf{P}}_{{k + 1}}-{\bf{P}}_{{i,k + 1}}$
		\end{algorithmic}
		$\ast$ \textbf{Return ${{\widetilde{\bf{X}}}_{k + 1}},{{{\bf{P}}}_{k + 1}},{{{\bf{P}}}_{i,k + 1}},{\bf{P}}_{d,k + 1}$}
	\end{spacing}	
\end{algorithm}

\subsection{Recursive Robot State Estimation With Accurate AprilTag Map}

\begin{figure*}[htbp]
	\centering
	\includegraphics[scale=0.90]{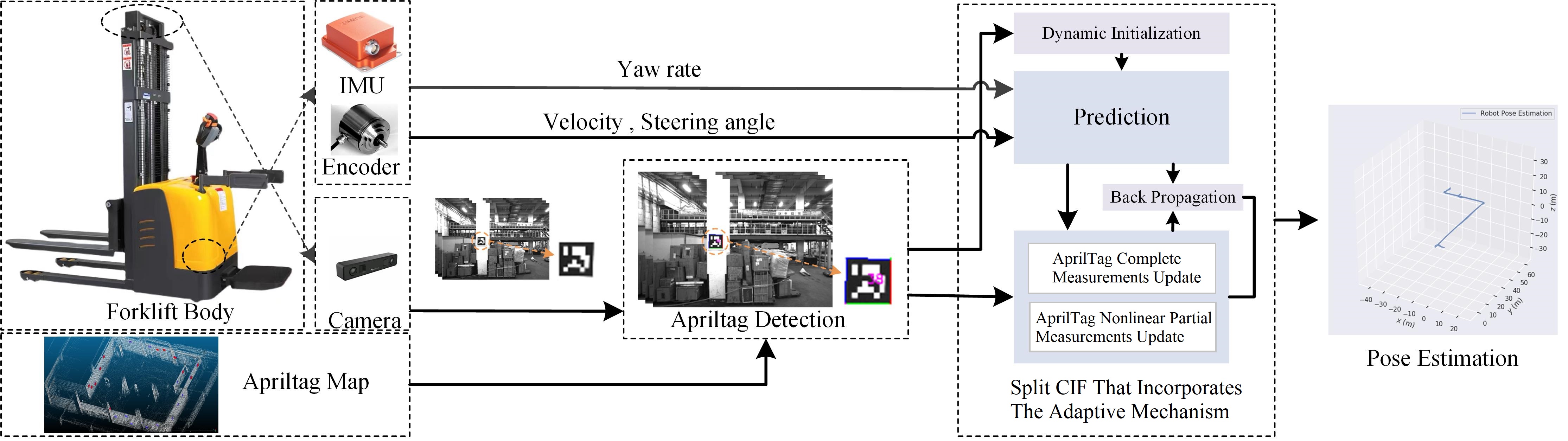} 	
	\caption{Overall system of recursive robot state estimation with accurate AprilTag map.}\label{Fig.2}
\end{figure*}

Once an accurate AprilTag map is available (see details in Section II(A)), visual localization is realized as recursive estimation that fuses AprilTag measurements (i.e. AprilTag detection results) and robot motion data, the overall system of which is illustrated in Fig. \ref{Fig.2}. The Split CIF that incorporates the measurement adaptive mechanism is used for fusion of AprilTag measurements and robot motion data.

The installation positions of the IMU (three-axis accelerometer and three-axis gyroscope), the encoder (linear encoder and rotary encoder) and the vision sensor are shown in the Fig. \ref{Fig.2}. The initialization can be realized by dynamic initialization mechanism, which can also initialize the pose in the process of pose estimation for solving the kidnapping. The prediction process is based on the derived forklift kinematic model with the motion data. The update process with complete AprilTag detection observation can prevent the damage of low quality observations via the presented observation noise model, and the extra partial measurement update process can use the hard-to-use observations for avoiding information loss. Moreover, for the existing latency situations in the image processing process, we integrate the back-projection into the recursive robot state estimation, which can effectively deal with the delay and guarantee the localization accuracy.

\subsubsection{Robot State Dynamic Initialization}

Before the recursive robot state estimation, the robot can obtain its initial global pose ${{\bf{X}}_{_{0}}}$ by the dynamic initialization, which can be calculated according to the first frame of AprilTag detection data matching with the map. And the state covariance ${{\bf{P}}_{_{0}}}$, process noise covariance ${{\bf{Q}}_{0}}$ can be set according to motion and visual sensors or real applications. Specifically, the ${{\bf{Q}}}$ can be set according to product parameters of IMU and odometer that input the control variables. In addition, the initial independent covariance ${{\bf{P}}_{_{i,0}}}$ can be the same as ${{\bf{P}}_{_{0}}}$, and dependent covariance ${{\bf{P}}_{_{d,0}}}$ can be set to $0$. And the independent part of observation noise covariance ${\bf{R}}_{{i,k+1}}$ can be replaced by the adaptive observation noise model in equation (2), and the initial dependent part can be regarded zero in the complete update observation process. Meanwhile, in partial update process, the initial independent part of split observation noise covariance can be set according to the statistic of the error of visual sensor detecting the distance in real applications, and the initial dependent part can be regarded zero. 

Moreover, when the kidnapping problem occurs, that is, at a certain moment in the navigation, a wrong pose with a large deviation may occur because of unexpected sudden collision, movement or human push, resulting in several consecutive estimated poses being discarded because they are inconsistent with this pose, and dynamic initialization mechanism will be enabled at this moment. Herein, the next measurement of the discarded measurements is used as the initial pose, and the state estimation process is restarted to ensure that the robot pose can be returned normally.

\subsubsection{Robot State Evolution (Prediction)}

The robot state evolution model $g(.)$ in Alg. \ref{Alg:ImplSplitCIF} reflects the relationship between the previous states ${{\bf{X}}_{k}}$ and its current state ${{\bf{X}}_{k+1}}$, which can be expressed in discrete form according to the forklift kinematic model:
\begin{equation}
\begin{small}
\left( \begin{array}{l}
\!\!{x_{k + 1}}\!\!\!\!\\
\!\!{y_{k + 1}}\!\!\!\!\\
\!\!{\theta _{k + 1}}\!\!\!\!
\end{array} \right) = \left( \begin{array}{l}
\!\!{x_{k}} + \Delta d\cos (\beta + {\theta _{k}} + \Delta \theta/2)\!\!\\
\!\!{y_{k}} + \Delta d\sin (\beta + {\theta _{k}} + \Delta \theta/2)\!\!\\
\quad\quad\quad\quad\!\!{\theta _{k}} + \Delta \theta\!\!
\end{array} \right) + \left( \begin{array}{l}
\!\!\!{\varepsilon _{{x}}}\!\!\!\!\\
\!\!\!{\varepsilon _{{y}}}\!\!\!\!\\
\!\!\!{\varepsilon _{{\theta}}}\!\!\!\!
\end{array} \right)
\end{small}
\end{equation}
where $x, y, \theta$ denote the robot pose. $\beta$ is the steering angle of front wheel, and the relationship of ${\theta _{k}}$ and $\beta$ denotes $sin({\theta _{k}}) /L = sin(\beta)/(v \cdot \Delta T)$. $\Delta \theta = \omega \cdot \Delta T$ and $\Delta d = v \cdot \Delta T$. $v$ and $\omega$ denote the robot velocity and yaw rate; $\Delta T$ denotes the time step; ${\varepsilon _{{x},{y},{\theta}}}$ are the process noise that arise from the encoder or wheels slipping.

\subsubsection{Robot State Update with AprilTag Complete Measurements}

The measurement adaptive mechanism that is incorporated into the Split CIF based visual localization solution consists in the adaptive AprilTag measurement noise model, details of which are formalized in Equation (2). The predicted robot state is updated with the complete pose measurement ${{\bf{Z}}_{k+1}} $ provided by the AprilTag detection module. Refer to the lines 1 to 7 of the part ``Update incorporating the measurement adaptive mechanism'' in Alg. \ref{Alg:ImplSplitCIF}. 

\subsubsection{Robot State Update with AprilTag Nonlinear Partial Measurements}

The measurement matrix ${\bf{H}}_{k+1}$ in above-mentioned measurement model is easily obtained. However, because of the existing extreme viewing distance and angle in image collection, the pixel detection error is magnified, leading to pose solving singularity problems, especially for the estimated angle. As a consequence, AprilTag complete measurements cannot be obtained, whereas AprilTag nonlinear partial measurements (namely AprilTag distance measurements) can only be obtained in such cases. Direct discarding of these AprilTag nonlinear partial measurements causes ``information waste'', because they can still contribute to robot state update. 

In the spirit of local linearization on which the extended Kalman filter (EKF) relies \cite{Li2022FARET_en} \cite{li2022FARET}, the Split CIF also enables update with nonlinear partial measurements. More specifically, suppose the warehouse robot has only a measurement of its distance to an AprilTag that has the globally-registered position \{$x_g, y_g$\} in the accurate AprilTag map established \textit{a priori}. The nonlinear partial measurement model is as follows, and we can locally linearize the measurement model about the prediction state \{${\tilde x}_{k + 1/k},{\tilde y}_{k + 1/k}$\} for the current period as follows:
\begin{equation}
\begin{small}
\begin{aligned}
{{\bf{Z}}_{k + 1}} &= h({{\bf{X}}_{k + 1}},k + 1) = \sqrt {{{({x_{k + 1}} - {x_g})}^2} + {{({y_{k + 1}} - {y_g})}^2}}\\
&\approx \sqrt {{{({{\tilde x}_{k + 1/k}} - {x_g})}^2} + {{({{\tilde y}_{k + 1/k}} - {y_g})}^2}}  \\&+ \frac{{({{\tilde x}_{k + 1/k}} - {x_g})({x_{k + 1}} - {{\tilde x}_{k + 1/k}})}}{{\sqrt {{{({{\tilde x}_{k + 1/k}} - {x_g})}^2} + {{({{\tilde y}_{k + 1/k}} - {y_g})}^2}} }}  \\&+ \frac{{({{\tilde y}_{k + 1/k}} - {y_g})({y_{k + 1}} - {{\tilde y}_{k + 1/k}})}}{{\sqrt {{{({{\tilde x}_{k + 1/k}} - {x_g})}^2} + {{({{\tilde y}_{k + 1/k}} - {y_g})}^2}} }}
\end{aligned}
\end{small}
\end{equation}

\begin{equation}
\begin{small}
\begin{aligned}
\begin{array}{l}
{{\bf{Z}}_{k + 1}} \approx {{\bf{D}}_{k + 1}} + {{\bf{C}}_{k + 1}}({x_{k + 1}} - {{\tilde x}_{k + 1/k}})  \\\quad\quad\quad\quad+ {{\bf{S}}_{k + 1}}({y_{k + 1}} - {{\tilde y}_{k + 1/k}})\\
\Leftrightarrow {{\bf{Z}}_{k + 1}} - {{\bf{D}}_{k + 1}} + {{\bf{C}}_{k + 1}}{{\tilde x}_{k + 1/k}}+ {{\bf{S}}_{k + 1}}{{\tilde y}_{k + 1/k}} \\\quad\;\;\approx \underbrace {{{\bf{C}}_{k + 1}}{x_{k + 1}} + {{\bf{S}}_{k + 1}}{y_{k + 1}}}_{{{{\bf{\widetilde Z}}}_{k + 1}}}\\
\Leftrightarrow {{{\bf{\widetilde Z}}}_{k + 1}} \approx \underbrace {\left[ {{{\bf{C}}_{k + 1}}\quad{{\bf{S}}_{k + 1}}\quad0} \right]}_{{\bf{H}}_{k + 1}}\left[ \begin{array}{l}
{x_{k + 1}}\\
{y_{k + 1}}\\
{\theta _{k + 1}}
\end{array} \right]
\end{array}	
\end{aligned}
\end{small}
\end{equation}

Where ${{{\bf{\widetilde Z}}}_{k + 1}} = {{\bf{Z}}_{k + 1}} - {{\bf{D}}_{k + 1}} + {{\bf{C}}_{_{k + 1}}}{{\tilde x}_{k + 1/k}} + {{\bf{S}}_{_{k + 1}}}{{\tilde y}_{k + 1/k}}$; and ${{\bf{H}}_{k + 1}}$ denotes the $\left[ {{{\bf{C}}_{k + 1}}\quad{{\bf{S}}_{k + 1}}\quad0} \right]$. Herein \begin{small}${{\bf{D}}_{k + 1}} = \sqrt {{{({{\tilde x}_{k + 1/k}} - {x_g})}^2} + {{({{\tilde y}_{k + 1/k}} - {y_g})}^2}}$\end{small}; \begin{small}${{\bf{C}}_{k + 1}} = {{({{\tilde x}_{k + 1/k}} - {x_g})}}/{{\sqrt {{{({{\tilde x}_{k + 1/k}} - {x_g})}^2} + {{({{\tilde y}_{k + 1/k}} - {y_g})}^2}} }}$\end{small} and \begin{small}${{\bf{S}}_{k + 1}} = {{({{\tilde y}_{k + 1/k}} - {y_g})}}/{{\sqrt {{{({{\tilde x}_{k + 1/k}} - {x_g})}^2} + {{({{\tilde y}_{k + 1/k}} - {y_g})}^2}} }}$\end{small}. 

Using this linearized distance measurement ${{\bf{\widetilde Z}}_{k+1}}$ instead of ${{\bf{Z}}_{k+1}}$ as shown in Alg. \ref{Alg:ImplSplitCIF}, we can still update the predicted robot state ${{\widetilde{\bf X}}_{k + 1/k}}$ with AprilTag nonlinear partial measurements via the proposed framework. 

\subsubsection{Back-projection}

In addition, our solution also consider the situation that the recursive estimation process received measurements are delayed since the image transmission and processing of the AprilTag is delayed for few seconds, which is common in practical applications and need to be resolved. The general solution that regard the error caused by delay as the random measurement error is not suitable for such large delay. Therefore, we apply the back-projection (BP) using motion data technology \cite{schafer2006motion, li2013cooperative} to compensate the delayed state estimate. Specifically, suppose the robot pose estimation (either via prediction or via update) at current time period and previous time periods are ${\bf X}_k$ and ${\bf X}_{k-1}, {\bf X}_{k-2}, ..., {\bf X}_{k-m}, ...$. For example, we can treat the motion data period as the estimate time period. Besides, store previous motion data ${\bf u}_t, {\bf u}_{k-1}, ...$ as well. We can store previous pose estimates and motion data in dynamically-adjusted queue (e.g. arrays) structures. Herein, the system prediction model is denoted as $g(\cdot)$, and it has \{${{\bf X}_k , {\bf P}_k}$\} = $g({{\bf X}_k , {\bf P}_k}, {\bf u}_k)$. Then, imagine at current time $k$, we receive a AprilTag measurement result whose timestamp is at (or near) time period $k-m$. In other words, we have an AprilTag measurement ${\bf Z}_{k-m}$. What the back-projection using motion data does is as follows:

a) Fuse (i.e. update) ${\bf X}_{k-m}$ with ${\bf Z}_{k-m}$ via the Split CIF as if ${\bf Z}_{k-m}$ was available at time $k-m$. Suppose the fused new result is \{${{\bf X}_{k-m,new}, {\bf P}_{k-m,new}}$\}. 

b) Back-project to update all pose estimates (together with their covariance) from time $k-m$ to $k$ in an iterative way as follows: 
\begin{equation}
\begin{small}
\begin{aligned}
\!\{{{\bf X}_{k-m+1,new},\! {\bf P}_{k-m+1,new}}\}\!\!=\!g(&{\bf X}_{k-m,new}, \\&{\bf P}_{k-m,new}, {\bf u}_{k-m+1})\\
\! \{{{\bf X}_{k-m+2,new},\! {\bf P}_{k-m+2,new}}\} \!\!=\!g(&{\bf X}_{k-m+1,new}, \\&{\bf P}_{k-m+1,new}, {\bf u}_{k-m+2})\\
  ...\quad&\\
\!  \{{{\bf X}_{k-1,new}, {\bf P}_{k-1,new}}\} \!\!=\!g(&{\bf X}_{k-2,new}, {\bf P}_{k-2,new}, {\bf u}_{k}) \\
\! \{{{\bf X}_{k,new}, {\bf P}_{k,new}}\} \!\!=\!g(&{\bf X}_{k-1,new}, {\bf P}_{k-1,new}, {\bf u}_{k}) 
\end{aligned}
\end{small}
\end{equation}

\section{Experimental Evaluation}

\subsection{Experiment Conditions}

To verify the performance of the proposed solution of split covariance intersection filter based visual localization with accurate AprilTag map, a comparative study of various experiments in real warehouse environments is performed. Here we implement an overall performance experiment using several comparative methods to evaluate our proposed system in terms of accuracy and robustness. In addition, we also provide more experimental results and analysis in the case of occurrence of kidnapping and image process information delay to demonstrate the performance of the proposed mechanisms. The several methods involved in the comparative study are as follows:

\textit{Pure AprilTag based visual localization (TagSLAM)}: The robot achieve the localization only rely on AprilTag detection module, without fusing the motion data.

\textit{Extended Kalman filter based visual localization that incorporates the measurement adaptive mechanism (EKF-Full)}: The robot resorts to the EKF for robot state update with AprilTag measurements, whereas implementation of other parts is the same to that of the proposed method.

\textit{Split Covariance Intersection Filter based visual localization without incorporating the measurement adaptive mechanism (SCIF-nonMA)}: The measurement adaptive mechanism is removed, whereas implementation of other parts is the same to that of the proposed method.

\textit{Split Covariance Intersection Filter based visual localization without updating AprilTag nonlinear partial measurements (SCIF-nonP)}: AprilTag nonlinear partial measurements are simply discarded without being used for robot state update, whereas implementation of other parts is the same to that of the proposed method.

\textit{Proposed solution of split covariance intersection filter based visual localization (SCIF-Full)}: The measurement adaptive mechanism is incorporated and AprilTag nonlinear partial measurements as well as AprilTag complete measurements are updated.

The hardware (IMU, encoder, camera) is installed on the warehouse forklift robot, as shown in Fig. {\ref{Fig.2}}, and has accurate calibration and hardware timestamp synchronization. Apriltags are installed on warehouse walls in the way such that they will not influence and will not be influenced by warehouse operations. The forklift robot drives safely according to the navigation destination in the factory. In order to ensure the synchronization of each sensor measurement, system time is added into the measured data as the time axis. And all the experiments are implemented on a laptop with 8 CPU of Intel Core i5-8265U 1.60 GHz.

\subsection{Performance of the overall system}

\begin{figure*}[htbp]
	\centering
	\includegraphics[scale=0.65]{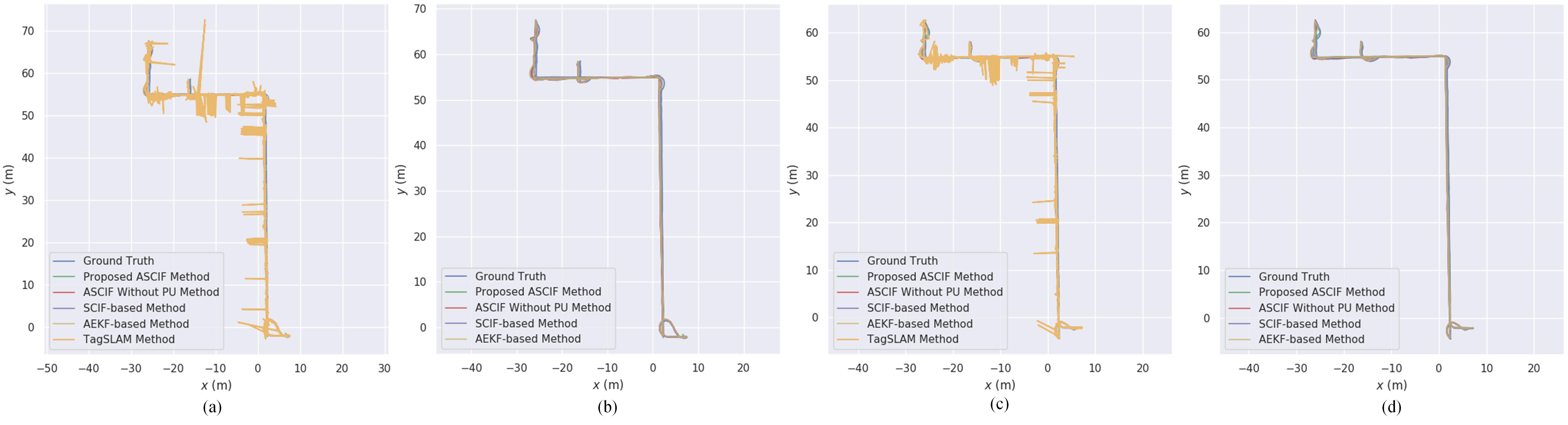}    		
	\caption{Test paths and localization trajectories for comparative methods. (a) Localization trajectories of all comparative methods in test path 1. (b) Trajectories of three comparative fusion localization methods in test path 1. (c) Localization trajectories of all comparative methods in test path 2. (d) Trajectories of three comparative fusion localization methods in test path 2.}\label{Fig.3}
\end{figure*}

In the experiments of testing the overall performance of proposed strategy, we compare the performance of forklift robot visual localization with the listed several methods executed simultaneously in two representative operation scenarios. The two test trajectories are shown in Fig. {\ref{Fig.3}}. Root mean square error (RMSE), Mean error (Mean) and standard error (STD) are used for localization error statistics and analysis, as shown in Table. {\ref{Tab.1}}.It can be seen that the performance of the Split CIF based fusion localization methods (including the SCIF-nonMA, the SCIF-nonP and the SCIF-Full) are better than the TagSLAM strategy in terms of accuracy. This verifies the advantages of the fusion localization and the proposed method can solve the problems of correlation and outlier efficiently for accurate localization. The various statistical error of the SCIF-Full visual localization method are smaller than the SCIF-nonMA visual localization method, which demonstrates that the measurement adaptive mechanism can improve the accuracy because it can adaptively adjust the weight of AprilTag measurements for robot state update. Besides, the SCIF-Full outperforms the SCIF-nonP, because it takes advantage of ``reasonable information'' contained in AprilTag nonlinear partial measurements instead of discarding all information directly. Moreover, the results in Table. {\ref{Tab.2}} demonstrate the robustness of the SCIF-Full in contrast with the pure AprilTag based visual localization method i.e. the TagSLAM.
	
Moreover, in order to further verify the SCIF-Full method for solving the correlation problem, we compare the proportional reduction of error of EKF-Full method and SCIF-Full-based method relative to error of TagSLAM method, as shown in Table. {\ref{Tab.3}}. It can be seen that the SCIF-Full can significantly reduce the error and have better performance compared with the EKF-Full that ignores temporal correlation among AprilTag measurements. This reflects the SCIF-Full can better handle potential temporal correlation among AprilTag measurements. Meanwhile, the experimental results in several different paths (the forklift robot is also tested on shorter or longer paths) demonstrate that the SCIF-Full visual localization method also has good and stable localization performance when applied to different scenarios.

\begin{table}[htbp]
	\centering
	\caption{Performance of localization for comparative methods}
	\label{Tab.1}
	\begin{center}
	\resizebox{8.6cm}{1.1cm}{
		\begin{tabular}{lcccccc}
			\toprule
			\multirow{1}{*}{\quad\quad Methods} & \multicolumn{3}{c}{ Test Path 1} &\multicolumn{3}{c}{ Test Path 2}\\
			\cmidrule(r){2-4} \cmidrule(r){5-7}
			&RMSE(m) &Mean(m) &STD(m)  &RMSE(m) &Mean(m) &STD(m) \\
			\midrule
			TagSLAM &0.79&0.55 &0.43 &0.58&0.50&0.39\\	
			\midrule
		 SCIF-nonMA &0.51&0.44&0.26 &0.47&0.42&0.25\\
			\midrule
			 SCIF-nonP &0.50&0.43 &0.25 &0.46&0.41&0.23\\
			\midrule
			SCIF-Full &0.45&0.39&0.21&0.40&0.37&0.20\\			
			\bottomrule
	\end{tabular}}
\end{center}
\end{table}
\begin{table}[htbp]
	\centering
	\caption{Reliability of localization for proposed method}
	\label{Tab.2}
	\begin{center}
		\resizebox{7.0cm}{0.65cm}{
			\begin{tabular}{lcc}
				\toprule
				\quad Success Rate of Method (\%)&Test Path 1 &Test Path 2 \\
				\midrule
				TagSLAM Success Rate &78\% &84\%\\
				\midrule
				SCIF-Full Success Rate&100 \%&100\%\\
				\bottomrule
		\end{tabular}}
	\end{center}
\end{table}
\begin{table}[htbp]
	\centering
	\caption{Proportional Reduction of Error for comparative methods}
	\label{Tab.3}
	\begin{center}
		\resizebox{8.6cm}{0.95cm}{
			\begin{tabular}{lcccccc}
				\toprule
			 \multicolumn{1}{c}{Proportional Reduction} & \multicolumn{2}{c}{ Test Path 1} &\multicolumn{2}{c}{ Test Path 2}\\
				\cmidrule(r){2-3} \cmidrule(r){4-5}
				 \quad\quad\quad\quad of Error (\%) & EKF-Full & SCIF-Full & EKF-Full & SCIF-Full \\
				\midrule
			 Proportional Reduction of RMSE &26\% &43\%&19\%&32\%\\
				\midrule
				 Proportional Reduction of Mean &12\% &30\%&10\%&26\%\\			
				\midrule
				 Proportional Reduction of STD&24\% &52\%&21\%&49\%\\
				\bottomrule
		\end{tabular}}
	\end{center}
\end{table}

\subsection{Case 1: Occurrence of kidnapping}

\begin{figure}[htbp]
	\centering
	\includegraphics[scale=0.43]{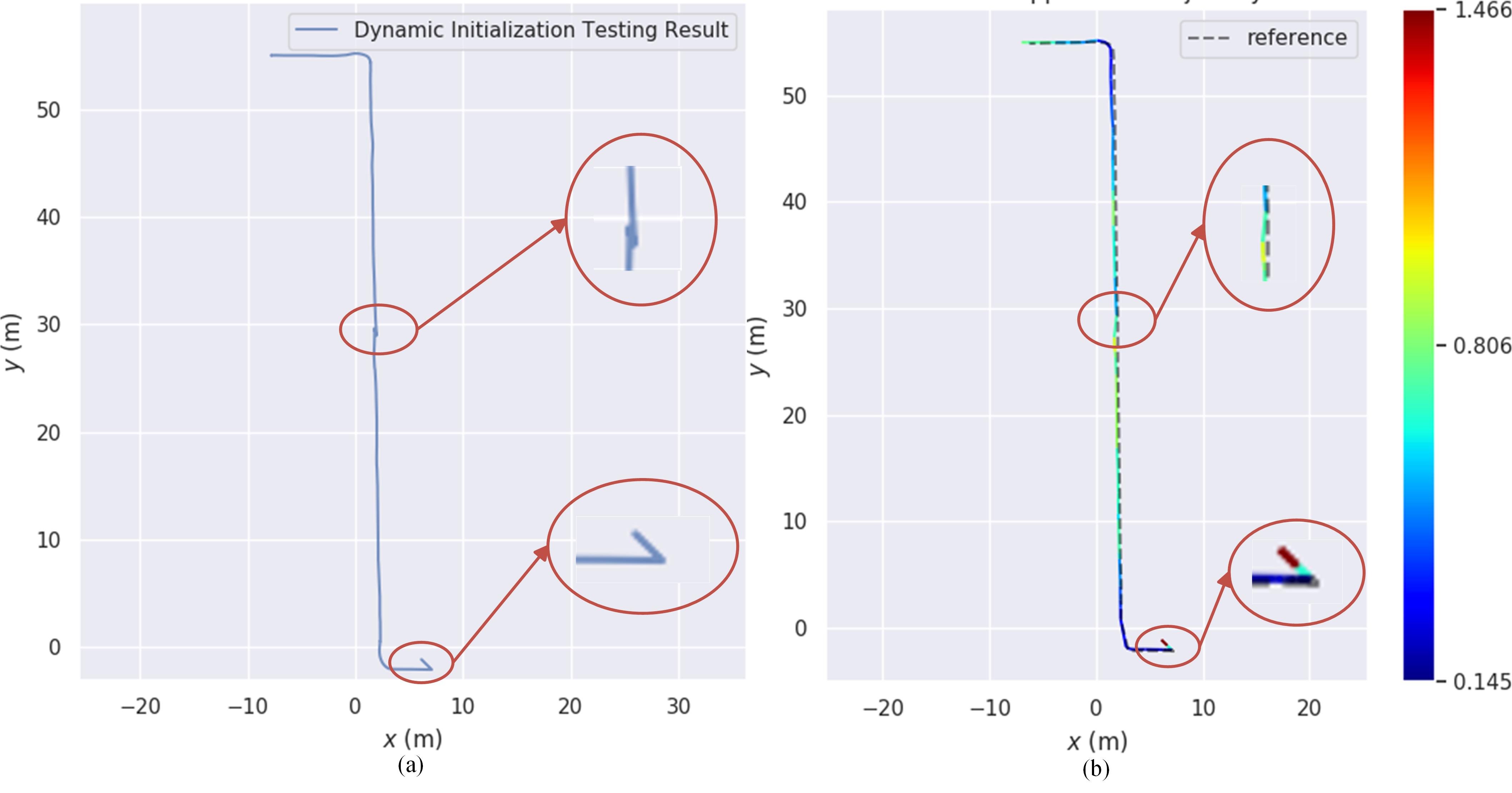}    		
	\caption{The trajectory (a) and error analysis (b) using proposed solution with dynamic initialization mechanism when kidnapping occurs. The two detailed figures in (a) and (b) show the trajectory changes and error change when kidnapping occurs, respectively. The color change in the two detail figures in (b) indicates that the error change from small to large and then decrease again.}\label{Fig.4}
\end{figure}

As discussed before, the pose estimation results may be kidnapped due to a wrong pose with a large deviation, which results in several consecutive estimated poses being discarded because they are inconsistent with this pose. Therefore, we presents the dynamic initialization mechanism, which can not only obtain the initial pose estimation but also solve this problem. To highlight the performances of dynamic initialization mechanism under the condition of occurrence of kidnapping more clearly, we have designed two kidnapping situations. When the forklift dynamically initialize the pose, we add 2 meters in one of the translation and subtract 2 meters to the other in the initial position, and we add 1.5 meter in the translations of the measurement at a certain moment in the robot normal navigation process to simulate the sudden movements of the robot leading to the wrong pose and potential kidnapping problem. The Fig. {\ref{Fig.4}} (a) and (b) show the partial localization trajectory using proposed solution with dynamic initialization mechanism and error analysis (we only show a short path here, the complete path have similar results), respectively. We observe that the wrong pose can affect the localization results unreliable, but soon returned to normal, see red circle marks in (a) of  figure and their corresponding error change statistics in (b). With the dynamic initialization mechanism, the kidnapping will not lead to several consecutive estimated poses being discarded due to inconsistent with the wrong pose and can maintaining a good performance in terms of robustness and accuracy.
\subsection{Case 2: Occurrence of delay}
About the occurrence of delay in the image transmission and processing process, it will result in the accuracy loss of the visual detection module, thereby reducing the fusion localization accuracy. Therefore, we presents the derived back-projection process in the recursive pose estimation (see details in Section II(D)), which can effectively deal with the latency and guarantee the localization accuracy. Specially, the comparative study with two test paths are shown in Table. {\ref{Tab.4}}. It can be seen that the various statistical error of proposed method are smaller than proposed method without adapting the back-projection technique, which demonstrates that not using back-projection to deal with the delay problem will cause a decrease in accuracy. The performed experimental results of whether to handle such delay in this way have proved the latency can be compensated. Meanwhile, the above technology for handling the delay have been implemented in other practical applications.

\begin{table}[htbp]
	\centering
	\caption{Localization error analysis in case of delay}
	\label{Tab.4}
	\begin{center}
		\resizebox{8.4cm}{0.75cm}{
			\begin{tabular}{lcccccc}
				\toprule
				\multirow{1}{*}{\quad\quad Methods} & \multicolumn{3}{c}{Test Path 1} &\multicolumn{3}{c}{Test Path 2}\\
				\cmidrule(r){2-4} \cmidrule(r){5-7}
				&RMSE(m) &Mean(m) &STD(m)  &RMSE(m) &Mean(m) &STD(m) \\
				\midrule
				SCIF-nonBP &0.47 &0.43&0.22 &0.44&0.42&0.21\\
				\midrule
				SCIF-Full &0.45&0.39&0.21&0.40&0.37&0.20\\
				\bottomrule
		\end{tabular}}
	\end{center}
\end{table}

\subsection{Discussion}

In addition, for the proposed solution of split covariance intersection filter based visual localization i.e. SCIF-Full, the measurement noise model involved in the measurement adaptive mechanism is established according to AprilTag measurement statistics. Specifically, we record the error of each measurement, the view angle and the view distance, and fit the relationship among them to establish a model. When discussing the effect of the number of installed Apriltags on the results, we also explore to appropriately reduce the number of Apriltags in straight or turns section of the test path. The results show that it has little overall impact on the proposed fusion localization method.

\section{CONCLUSIONS}

Split covariance intersection filter based visual localization with accurate AprilTag map has been proposed, aiming at providing a reliable and commercially-competitive solution for warehouse applications. As highlights in the proposed solution, first, an accurate AprilTag map is established with the help of a LiDAR-based SLAM system (namely the unique mapping robot). There would be no cost concern because the cost of the mapping robot is shared by a large amount of operating robots that can all benefit from the established accurate AprilTag map. Second, once the accurate AprilTag map is available, for each of the large amount of operating robots, visual localization is realized as recursive estimation that fuses AprilTag measurements and robot motion data, taking advantage of the split covariance intersection filter that can handle temporal correlation among AprilTag measurements and handle AprilTag nonlinear partial measurements in the spirit of local linearization as well. Besides, each operating robot incorporates a measurement adaptive mechanism to handle outliers in AprilTag measurements and adopts a dynamic initialization mechanism to address the kidnapping problem. A comparative study of various experiments in real warehouse environments demonstrate the potential and advantage of split covariance intersection filter based visual localization with accurate AprilTag map.

As communication devices with highly-qualified performance can be commercially deployed more and more nowadays, for future extensions in practice, multiple warehouse robots that operate in the spirit of cooperative visual localization may be studied.

\bibliographystyle{IEEEtran}
\bibliography{TagLocalization}

\begin{thebibliography}{10}
\providecommand{\url}[1]{#1}
\csname url@samestyle\endcsname
\providecommand{\newblock}{\relax}
\providecommand{\bibinfo}[2]{#2}
\providecommand{\BIBentrySTDinterwordspacing}{\spaceskip=0pt\relax}
\providecommand{\BIBentryALTinterwordstretchfactor}{4}
\providecommand{\BIBentryALTinterwordspacing}{\spaceskip=\fontdimen2\font plus
\BIBentryALTinterwordstretchfactor\fontdimen3\font minus
  \fontdimen4\font\relax}
\providecommand{\BIBforeignlanguage}[2]{{%
\expandafter\ifx\csname l@#1\endcsname\relax
\typeout{** WARNING: IEEEtran.bst: No hyphenation pattern has been}%
\typeout{** loaded for the language `#1'. Using the pattern for}%
\typeout{** the default language instead.}%
\else
\language=\csname l@#1\endcsname
\fi
#2}}
\providecommand{\BIBdecl}{\relax}
\BIBdecl

\bibitem{durrant2006simultaneous}
H.~Durrant-Whyte and T.~Bailey, ``Simultaneous localization and mapping: {P}art
  {I},'' \emph{IEEE robotics \& automation magazine}, vol.~13, no.~2, pp.
  99--110, 2006.

\bibitem{bailey2006simultaneous}
T.~Bailey and H.~Durrant-Whyte, ``Simultaneous localization and mapping
  ({SLAM}): {P}art {II},'' \emph{IEEE robotics \& automation magazine},
  vol.~13, no.~3, pp. 108--117, 2006.

\bibitem{vasiljevic2016high}
G.~Vasiljevi{\'c}, D.~Mikli{\'c}, I.~Draganjac, Z.~Kova{\v{c}}i{\'c}, and
  P.~Lista, ``High-accuracy vehicle localization for autonomous warehousing,''
  \emph{Robotics and Computer-Integrated Manufacturing}, vol.~42, pp. 1--16,
  2016.

\bibitem{bresson2017simultaneous}
G.~Bresson, Z.~Alsayed, L.~Yu, and S.~Glaser, ``Simultaneous localization and
  mapping: {A} survey of current trends in autonomous driving,'' \emph{IEEE
  Transactions on Intelligent Vehicles}, vol.~2, no.~3, pp. 194--220, 2017.

\bibitem{yu2013visual}
H.-H. Yu, H.-W. Hsieh, Y.-K. Tasi, Z.-H. Ou, Y.-S. Huang, and T.~Fukuda,
  ``Visual localization for mobile robots based on composite map,''
  \emph{Journal of Robotics and Mechatronics}, vol.~25, no.~1, pp. 25--37,
  2013.

\bibitem{li2019deep}
C.~Li, S.~Wang, Y.~Zhuang, and F.~Yan, ``Deep sensor fusion between {2D} laser
  scanner and {IMU} for mobile robot localization,'' \emph{IEEE Sensors
  Journal}, vol.~21, no.~6, pp. 8501--8509, 2019.

\bibitem{xu2021indoor}
X.~Xu, F.~Pang, Y.~Ran, Y.~Bai, L.~Zhang, Z.~Tan, C.~Wei, and M.~Luo, ``An
  indoor mobile robot positioning algorithm based on adaptive federated
  {K}alman filter,'' \emph{IEEE Sensors Journal}, vol.~21, no.~20, pp.
  23\,098--23\,107, 2021.

\bibitem{Li2024IV}
Z.~Ying and H.~Li, ``I{MM-SLAMMOT}: {T}ightly-coupled {SLAM} and {IMM}-based
  multi-object tracking,'' \emph{IEEE Transactions on Intelligent Vehicles},
  vol.~9, no.~2, pp. 3964--3974, 2024.

\bibitem{mur2015orb}
R.~Mur-Artal, J.~M.~M. Montiel, and J.~D. Tardos, ``O{RB-SLAM}: a versatile and
  accurate monocular {SLAM} system,'' \emph{IEEE transactions on robotics},
  vol.~31, no.~5, pp. 1147--1163, 2015.

\bibitem{sumikura2019openvslam}
S.~Sumikura, M.~Shibuya, and K.~Sakurada, ``Open{VSLAM}: {A} versatile visual
  {SLAM} framework,'' in \emph{Proceedings of the 27th ACM International
  Conference on Multimedia}, 2019, pp. 2292--2295.

\bibitem{campos2021orb}
C.~Campos, R.~Elvira, J.~J.~G. Rodr{\'\i}guez, J.~M. Montiel, and J.~D.
  Tard{\'o}s, ``O{RB-SLAM}3: {A}n accurate open-source library for visual,
  visual--inertial, and multimap {SLAM},'' \emph{IEEE Transactions on
  Robotics}, vol.~37, no.~6, pp. 1874--1890, 2021.

\bibitem{bloesch2015robust}
M.~Bloesch, S.~Omari, M.~Hutter, and R.~Siegwart, ``Robust visual inertial
  odometry using a direct {EKF}-based approach,'' in \emph{2015 IEEE/RSJ
  international conference on intelligent robots and systems (IROS)}.\hskip 1em
  plus 0.5em minus 0.4em\relax IEEE, 2015, pp. 298--304.

\bibitem{qin2018vins}
T.~Qin, P.~Li, and S.~Shen, ``V{INS}-mono: {A} robust and versatile monocular
  visual-inertial state estimator,'' \emph{IEEE Transactions on Robotics},
  vol.~34, no.~4, pp. 1004--1020, 2018.

\bibitem{Li2024arXivVisualSLAMMOT}
P.~Tian and H.~Li, ``Visual {SLAMMOT} considering multiple motion models,''
  \emph{arXiv}, 2024.

\bibitem{zheng2008mr}
R.~Zheng and K.~Yuan, ``M{R} code for indoor robot self-localization,'' in
  \emph{2008 7th World Congress on Intelligent Control and Automation}.\hskip
  1em plus 0.5em minus 0.4em\relax IEEE, 2008, pp. 7449--7454.

\bibitem{fourmy2019absolute}
M.~Fourmy, D.~Atchuthan, N.~Mansard, J.~Sola, and T.~Flayols, ``Absolute
  humanoid localization and mapping based on {IMU} {L}ie group and fiducial
  markers,'' in \emph{2019 IEEE-RAS 19th International Conference on Humanoid
  Robots (Humanoids)}.\hskip 1em plus 0.5em minus 0.4em\relax IEEE, 2019, pp.
  237--243.

\bibitem{olson2011apriltag}
E.~Olson, ``Apriltag: {A} robust and flexible visual fiducial system,'' in
  \emph{2011 IEEE international conference on robotics and automation}.\hskip
  1em plus 0.5em minus 0.4em\relax IEEE, 2011, pp. 3400--3407.

\bibitem{wang2016apriltag}
J.~Wang and E.~Olson, ``Apriltag 2: {E}fficient and robust fiducial
  detection,'' in \emph{2016 IEEE/RSJ International Conference on Intelligent
  Robots and Systems (IROS)}.\hskip 1em plus 0.5em minus 0.4em\relax IEEE,
  2016, pp. 4193--4198.

\bibitem{pfrommer2019tagslam}
B.~Pfrommer and K.~Daniilidis, ``Tag{SLAM}: {R}obust slam with fiducial
  markers,'' \emph{arXiv preprint arXiv:1910.00679}, 2019.

\bibitem{kayhani2019improved}
N.~Kayhani, A.~Heins, W.~Zhao, M.~Nahangi, B.~McCabe, and A.~P. Schoelligb,
  ``Improved tag-based indoor localization of {UAV}s using extended {K}alman
  filter,'' in \emph{Proceedings of the ISARC. International Symposium on
  Automation and Robotics in Construction, Banff, AB, Canada}, 2019, pp.
  21--24.

\bibitem{yu2021indoor}
L.~Yu, M.~Li, and G.~Pan, ``Indoor localization based on fusion of {A}priltag
  and adaptive monte carlo,'' in \emph{2021 IEEE 5th Information Technology,
  Networking, Electronic and Automation Control Conference (ITNEC)},
  vol.~5.\hskip 1em plus 0.5em minus 0.4em\relax IEEE, 2021, pp. 464--468.

\bibitem{hanley2021impact}
D.~Hanley, A.~S.~D. De~Oliveira, X.~Zhang, D.~H. Kim, Y.~Wei, and T.~Bretl,
  ``The impact of height on indoor positioning with magnetic fields,''
  \emph{IEEE Transactions on Instrumentation and Measurement}, vol.~70, pp.
  1--19, 2021.

\bibitem{li2014mobile}
I.-H. Li, M.-C. Chen, W.-Y. Wang, S.-F. Su, and T.-W. Lai, ``Mobile robot
  self-localization system using single webcam distance measurement technology
  in indoor environments,'' \emph{Sensors}, vol.~14, no.~2, pp. 2089--2109,
  2014.

\bibitem{ullah2020simultaneous}
I.~Ullah, X.~Su, X.~Zhang, and D.~Choi, ``Simultaneous localization and mapping
  based on {K}alman filter and extended {K}alman filter,'' \emph{Wireless
  Communications and Mobile Computing}, vol. 2020, 2020.

\bibitem{Li2022FARET_en}
\begin{CJK}{UTF8}{gbsn}李颢\end{CJK},
  \emph{\begin{CJK}{UTF8}{gbsn}迭代估计理论基础与应用（英文版）\end{CJK}}.\hskip
  1em plus 0.5em minus 0.4em\relax
  \begin{CJK}{UTF8}{gbsn}上海交通大学出版社\end{CJK}, 2022.

\bibitem{li2022FARET}
H.~Li, \emph{Fundamentals and applications of recursive estimation
  theory}.\hskip 1em plus 0.5em minus 0.4em\relax Shanghai Jiao Tong University
  Press, 2022.

\bibitem{Julier2001}
S.~Julier and J.~Uhlmann, ``General decentralized data fusion with covariance
  intersection (ci),'' \emph{Handbook of Data Fusion}, 2001.

\bibitem{li2013split}
H.~Li, F.~Nashashibi, and M.~Yang, ``Split covariance intersection filter:
  {T}heory and its application to vehicle localization,'' \emph{IEEE
  Transactions on Intelligent Transportation Systems}, vol.~14, no.~4, pp.
  1860--1871, 2013.

\bibitem{Cros2025}
C.~Cros, P.~Amblard, C.~Prieur, and J.~D. Rocha, ``Revisiting split covariance
  intersection: Correlated components and optimality,'' \emph{IEEE Transactions
  on Automatic Control}, vol.~70, no.~7, pp. 4593--4607, 2025.

\bibitem{li2013cooperative}
H.~Li and F.~Nashashibi, ``Cooperative multi-vehicle localization using split
  covariance intersection filter,'' \emph{IEEE Intelligent transportation
  systems magazine}, vol.~5, no.~2, pp. 33--44, 2013.

\bibitem{Li2013d}
H.~Li, F.~Nashashibi, B.~Lefaudeux, and E.~Pollard, ``Track-to-track fusion
  using split covariance intersection filter-information matrix filter
  (scif-imf) for vehicle surrounding environment perception,'' in \emph{IEEE
  International Conference on Intelligent Transportation Systems}, 2013, pp.
  1430--1435.

\bibitem{Wanasinghe2014}
T.~R. Wanasinghe, G.~K.~I. Mann, and R.~G. Gosine, ``Decentralized cooperative
  localization for heterogeneous multi-robot system using split covariance
  intersection filter,'' in \emph{Canadian Conference on Computer and Robot
  Vision}, 2014, pp. 167--174.

\bibitem{Pierre2018}
C.~Pierre, R.~Chapuis, R.~Aufrère, J.~Laneurit, and C.~Debain, ``Range-only
  based cooperative localization for mobile robots,'' in \emph{International
  Conference on Information Fusion}, 2018, pp. 1933--1939.

\bibitem{Chen2020}
X.~Chen, M.~Yang, W.~Yuan, H.~Li, and C.~Wang, ``Split covariance intersection
  filter based front-vehicle track estimation for vehicle platooning without
  communication,'' in \emph{IEEE Intelligent Vehicles Symposium}, 2020, pp.
  1510--1515.

\bibitem{fang2022inertial}
S.~Fang, H.~Li, M.~Yang, and Z.~Wang, ``Inertial navigation system based
  vehicle temporal relative localization with split covariance intersection
  filter,'' \emph{IEEE Robotics and Automation Letters}, vol.~7, no.~2, pp.
  5270--5277, 2022.

\bibitem{Allig2022}
C.~Allig and G.~Wanielik, ``Unequal dimension track-to-track fusion approaches
  using covariance intersection,'' \emph{IEEE Transactions on Intelligent
  Transportation Systems}, vol.~23, no.~6, pp. 5881--5886, 2022.

\bibitem{Li2022TITS}
S.~Fang, H.~Li, and M.~Yang, ``Lidar slam based multivehicle cooperative
  localization using iterated split cif,'' \emph{IEEE Transactions on
  Intelligent Transportation Systems}, vol.~23, no.~11, pp. 21\,137--21\,147,
  2022.

\bibitem{zair2016outlier}
S.~Zair, L.~H{\'e}garat-Mascle, E.~Seignez \emph{et~al.}, ``Outlier detection
  in {GNSS} pseudo-range/{D}oppler measurements for robust localization,''
  \emph{Sensors}, vol.~16, no.~4, p. 580, 2016.

\bibitem{bai2018robust}
F.~Bai, T.~Vidal-Calleja, and S.~Huang, ``Robust incremental {SLAM} under
  constrained optimization formulation,'' \emph{IEEE Robotics and Automation
  Letters}, vol.~3, no.~2, pp. 1207--1214, 2018.

\bibitem{shojaei2011experimental}
K.~Shojaei and A.~M. Shahri, ``Experimental study of iterated {K}alman filters
  for simultaneous localization and mapping of autonomous mobile robots,''
  \emph{Journal of Intelligent \& Robotic Systems}, vol.~63, no. 3-4, pp.
  575--594, 2011.

\bibitem{zhao2016robust}
J.~Zhao, M.~Netto, and L.~Mili, ``A robust iterated extended {K}alman filter
  for power system dynamic state estimation,'' \emph{IEEE Transactions on Power
  Systems}, vol.~32, no.~4, pp. 3205--3216, 2016.

\bibitem{huang2017new}
Y.~Huang, Y.~Zhang, B.~Xu, Z.~Wu, and J.~A. Chambers, ``A new adaptive extended
  {K}alman filter for cooperative localization,'' \emph{IEEE Transactions on
  Aerospace and Electronic Systems}, vol.~54, no.~1, pp. 353--368, 2017.

\bibitem{wang2019adaptive}
Y.~Wang, Y.~Sun, V.~Dinavahi, S.~Cao, and D.~Hou, ``Adaptive robust cubature
  {K}alman filter for power system dynamic state estimation against outliers,''
  \emph{IEEE Access}, vol.~7, pp. 105\,872--105\,881, 2019.

\bibitem{shan2018lego}
T.~Shan and B.~Englot, ``{L}e{GO}-{LOAM}: Lightweight and ground-optimized
  lidar odometry and mapping on variable terrain,'' in \emph{2018 IEEE/RSJ
  International Conference on Intelligent Robots and Systems (IROS)}.\hskip 1em
  plus 0.5em minus 0.4em\relax IEEE, 2018, pp. 4758--4765.

\bibitem{Li2013calib}
H.~Li and F.~Nashashibi, ``Comprehensive extrinsic calibration of a camera and
  a 2d laser scanner for a ground vehicle,'' \emph{INRIA Tech Report}, vol.
  RT-438, 2013.

\bibitem{dellaert2012factor}
F.~Dellaert, ``Factor graphs and {GTSAM}: {A} hands-on introduction,'' Georgia
  Institute of Technology, Tech. Rep., 2012.

\bibitem{schafer2006motion}
D.~Schafer, J.~Borgert, V.~Rasche, and M.~Grass, ``Motion-compensated and gated
  cone beam filtered back-projection for 3-{D} rotational {X}-ray
  angiography,'' \emph{IEEE transactions on medical imaging}, vol.~25, no.~7,
  pp. 898--906, 2006.

\end{thebibliography}

\end{document}